\documentclass[conference]{IEEEtran}
\IEEEoverridecommandlockouts
\usepackage{amsmath,amssymb,amsfonts}
\usepackage{algorithmic}
\usepackage{graphicx}
\usepackage{textcomp}
\usepackage{xcolor}
\usepackage{booktabs}
\usepackage[numbers,sort&compress]{natbib}
\usepackage{balance}
\usepackage{tikz}
\usepackage{hyperref}
\usepackage{makecell}
\usepackage{pifont}
\usepackage[capitalise,nameinlink]{cleveref}
\usetikzlibrary{arrows.meta, fit, backgrounds, positioning}
\usepackage{float}
\usepackage[most]{tcolorbox}
\newtcolorbox{defbox}{
  colback=black!3, colframe=black!45,
  boxrule=0.4pt, arc=1pt,
  left=5pt, right=5pt, top=4pt, bottom=4pt,
  boxsep=0pt
}

\newtheorem{definition}{Definition}[section]
\newtheorem{proposition}[definition]{Proposition}

\def\BibTeX{{\rm B\kern-.05em{\sc i\kern-.025em b}\kern-.08em
    T\kern-.1667em\lower.7ex\hbox{E}\kern-.125emX}}
\begin{document}

\title{Inversion-Free High-Dimensional Counterfactual Inference}


\author{
\IEEEauthorblockN{Saketh Vishnubhatla\IEEEauthorrefmark{1},
Pingchuan Ma\IEEEauthorrefmark{1},
Shu Wan\IEEEauthorrefmark{1},
Alimohammad Beigi\IEEEauthorrefmark{1},
Andre Harrison\IEEEauthorrefmark{2},
Adrienne Raglin\IEEEauthorrefmark{2},
Huan Liu\IEEEauthorrefmark{1}}
\IEEEauthorblockA{\IEEEauthorrefmark{1}School of Computing and Augmented
Intelligence, Arizona State University, Tempe, AZ 85281, USA\\
\{svishnu6, pingchua, swan16, abeigi, huanliu\}@asu.edu}
\IEEEauthorblockA{\IEEEauthorrefmark{2}DEVCOM Army Research Laboratory,
Adelphi, MD 20783, USA\\
\{andre.v.harrison2.civ, adrienne.raglin2.civ\}@army.mil}
}

\maketitle

\begin{abstract}
Counterfactual inference is the task of answering hypothetical “what-if” queries about a unit, given an observational dataset, modeling assumptions, and a causal graph. This enables a modeler to reason about hypothetical scenarios that cannot otherwise be answered without assumptions or a causal graph. The simpler case involves a causal graph with scalar 1-D variables that denote different attributes in the data-generation process. While the literature is clear in the 1-D case, extending this to settings with high-dimensional nodes (e.g., images) remains challenging. Conventionally, in the literature, methods recover the latent state from the factual image (inversion), generate counterfactual attributes from the graph, and render the counterfactual image from the latent state using the updated attributes. In our work, we investigate the necessity of the inversion step in computing high-dimensional counterfactuals. We first discuss moving from the scalar 1-D case to the higher-dimensional case. Secondly, we discuss the relevance of the inversion step and evaluate an inversion-free counterpart repurposed from the image-editing literature. Finally, on image datasets such as CelebA and ADNI, we benchmark our approach using two measures: namely, effectiveness and identity preservation. While effectiveness is a measure of how faithfully the counterfactual attribute is rendered, identity preservation measures whether the same unit (image) is being rendered, both of which are crucial for high-dimensional counterfactual inference. We present the tradeoffs of using inversion-based and inversion-free approaches in our work.    
\end{abstract}


\section{Introduction}
\label{sec:introduction}
Answering hypothetical questions about individual cases are common in data analysis and
hard to answer without assumptions. Would this patient's chances of survival looked different had the disease progressed further? This is a counterfactual query, and Pearl's abduction--action--prediction procedure \cite{pearl2009causality} answers the query when the variables concerned are scalars, under some assumptions that are by now well understood. However, most of the variables of interest are not single-dimensional scalars. A brain scan or a photograph carries several attributes at once, many of them causally related, and answering a counterfactual query about such a record means producing the record itself rather than a number. This has become feasible, with generative models capable of rendering an image conditioned on a requested set of attributes.

The pipelines that answer such queries share a common structure. A causal model over the attributes, a generative backbone that renders the image, and between them sits a step that recovers a latent state from the factual image, intended to make the result depict the
same unit rather than a fresh image carrying the requested attributes. This state is taken from a learned encoder~\citep{pawlowski2020deep,ribeiro2023high} or from running the backbone backward to a noise map~\citep{sanchez2022diffusion, rasal2025diffusion, tong2025causal}, and the step is also referred to as ``abduction'', taking its name after the procedure of inverting a structural equation the modeler has fitted in the standard 1-D case. The name is warranted when the image mechanism is trained as part of the causal model, and identification in high dimensions has been established on that basis~\citep{ribeiro2025identifiability}, under conditions that are demanding in practice. We investigate the role of this inversion step in computing counterfactuals in the higher dimensional case.

Our contributions are as follows:

\begin{figure*}[h]
\centering
\includegraphics[width=0.9\textwidth]{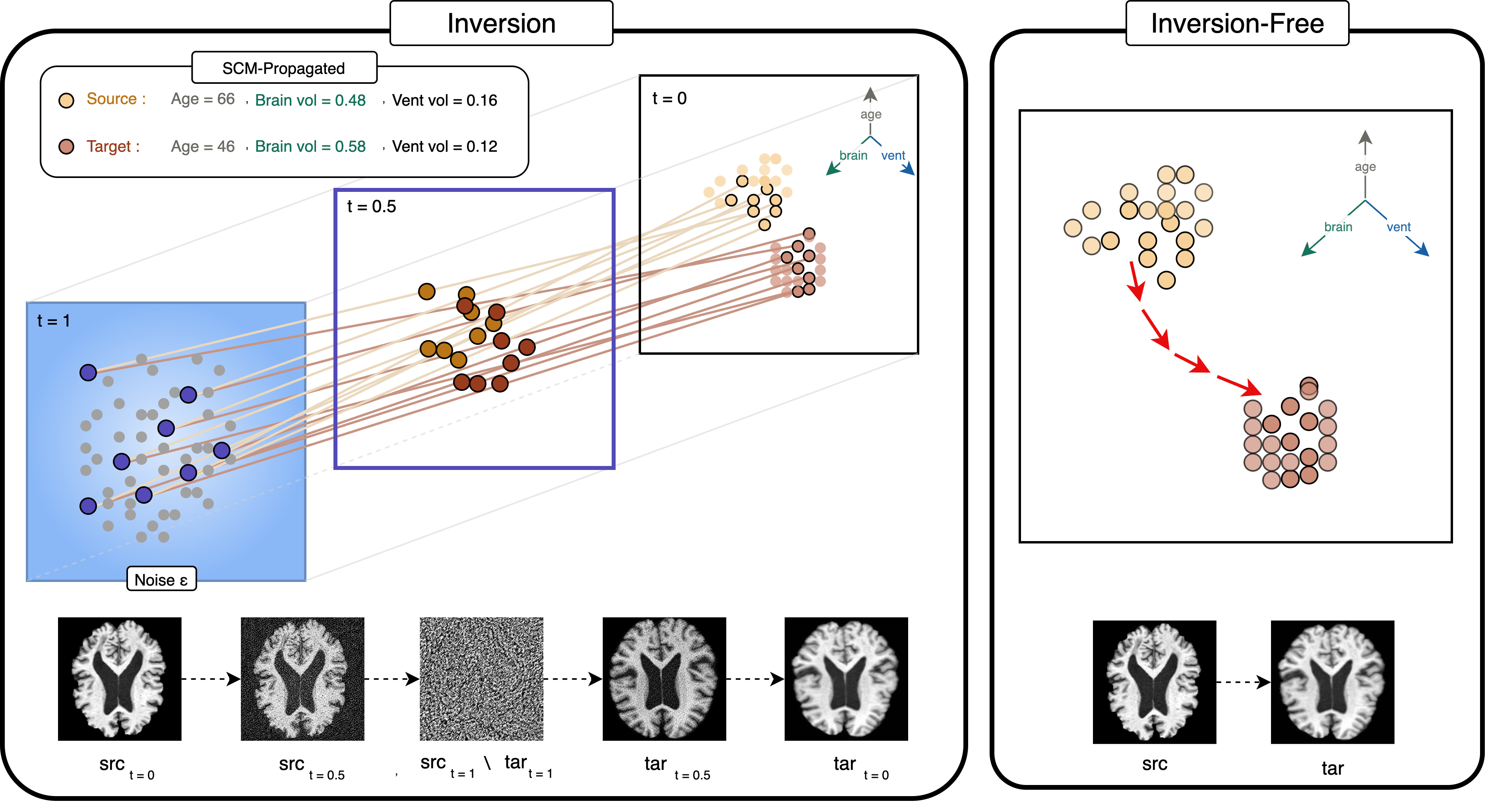}
\caption{The two transports, both given the same counterfactual attributes
obtained by propagating the antecedent on age through the structural
equations. \textbf{Left:} inversion carries the factual to noise under the
source condition and regenerates under the target. \textbf{Right:} the
inversion-free map moves the factual directly, by accumulated velocity
differences between the two conditions, always staying on the image manifold.}
\label{fig:proposal}
\end{figure*}

\begin{itemize}
  \item Analysing the issues in moving from 1-D counterfactual inference to higher dimensional counterfactual inference, discussing approaches in the literature and identification arguments involved to tackle the higher dimensional case.
  \item Presenting the role of inversion step, conventionally used by most works in computing higher-dimensional counterfactuals, and proposing an inversion-free pipeline that does not generate the latent during the edit.
  \item Testing the inversion-based approaches against our proposed inversion-free approach on image datasets like CelebA, a corpus of celebrity images and ADNI, consisting of brain MRI scans. We provide tradeoffs on two measures namely effectiveness and identity preservation.
\end{itemize}

\section{Related Work}
We classify works on counterfactual inference into three categories: starting with works relevant to traditional counterfactual inference, literature dealing with using deep networks for the task and the body of work relevant to counterfactual queries over images.

\paragraph{Counterfactual Inference}
Counterfactual queries are posed within the Structural Causal Model framework \citep{pearl2009causality}, to answer a query about a specific unit through the abduction--action--prediction procedure and, with the axiomatic characterization of \citet{galles1998axiomatic}. The difficulty lies in abduction, and it is resolved by restricting the functional form of the mechanism, a restriction that the literature has progressively weakened. Additive noise models fix how the exogenous term enters, so that it is recovered as a residual while the dependence on the parents remains nonlinear \citep{hoyer2008nonlinear}. Post-nonlinear models relax additivity itself by composing the additive form with an invertible output nonlinearity, which is inverted before the residual is taken \citep{zhang2012identifiability}. \citet{nasr2023counterfactual} unify these under bijective generation mechanisms and establish counterfactual identifiability for three causal structures, in the Markovian case relevant here,
strict monotonicity in the exogenous argument. The other two cases admit unobserved confounding, identifying the mechanism through an instrumental variable or a backdoor set in place of the Markovian assumption. Recently another approach by Von Kuglegen et al.~\cite{von2023backtracking} popularised \textit{backtracking counterfactuals}. This paradigm allows for the same causal graph while computing the counterfactual, breaking away from the traditional interventionist account.

\paragraph{Deep Networks for Counterfactual Inference}
Scaling causal inference to high-dimensional data has led to the development of Deep Structural Causal Models (DSCMs)~\cite{pawlowski2020deep}, which leverage normalizing flows and VAEs to parameterize exogenous noise. Neural Causal Models~\cite{xia2023neural} further extend this by providing frameworks for identification and estimation in complex, non-linear environments. Recent research has specifically addressed the problem of \textit{Deep Backtracking}~\cite{kladny2024deep}, proposing methods that ensure counterfactual explanations remain "causally compliant" by allowing background variables to be updated. Other developments include handling categorical background variables~\cite{de2022deep} and ensuring end-to-end consistency for counterfactual fairness~\cite{ma2026consistent}, ensuring that deep estimators do not introduce spurious correlations during the abduction phase.

\paragraph{Image Editing}
In the visual domain, counterfactual editing seeks to modify specific image attributes while preserving the underlying causal structure. Early generative approaches often conflated causal dynamics with image rendering, but recent work on Augmented SCMs (ASCM)~\cite{pan2024counterfactual} has formalized the separation of mixing functions from causal mechanisms. High-fidelity image counterfactuals have been achieved using probabilistic causal models~\cite{ribeiro2023high}, which utilize latent space abduction to ensure realistic reconstructions. The emergence of diffusion-based models has further pushed the state-of-the-art. Diffusion Causal Models~\cite{sanchez2022diffusion} allow for fine-grained counterfactual estimation by treating the reverse diffusion process as a causal mechanism. More recently, semantic abduction techniques~\cite{rasal2025diffusion} have improved the alignment between textual prompts and causal interventions, enabling diffusion models to generate images that are not only visually convincing but also causally consistent with complex counterfactual queries. A parallel line in the vision literature avoids the round trip altogether,
editing along the generative path~\cite{Kulikov_2025_ICCV, kim2026flowalign}.
These methods carry no causal model over the attributes, but supply the
transports compared here against inversion.

\section{Preliminaries}
\label{sec:preliminaries}
\label{sec:prelim}

Lowercase letters denote scalars, lowercase boldface vectors, uppercase
boldface matrices, and calligraphic letters sets and tuples that are not random variables. Plain uppercase
letters denote random variables and their lowercase counterparts realized
values. We write $P(\cdot)$ for probability distributions and
$\mathrm{do}(\cdot)$ for Pearl's intervention operator. The remainder of this section focuses on structural causal models, counterfactuals, the notion of counterfactual identifiability and some basics of flow-based generative models that we later use.

\paragraph{Structural Causal Models}
A Structural Causal Model (SCM) is a tuple 
$\mathcal{M} = \langle U, V, \mathcal{F}, P(U) \rangle$, where $U$ is a set 
of exogenous noise variables with prior $P(U)$, $V$ is a set of endogenous 
variables, and $\mathcal{F} = \{f_i\}$ is a set of structural equations 
$v_i = f_i(\mathrm{pa}_i, u_i)$ mapping parents and noise to each 
variable~\citep{pearl2009causality}. Each SCM induces a causal diagram 
$\mathcal{G}$ in which directed edges encode direct causal influence and 
bidirected edges encode shared exogenous noise. An intervention 
$\mathrm{do}(v_i = v^*)$ replaces $f_i$ with the constant $v^*$, yielding a 
submodel $\mathcal{M}_{v_i = v^*}$ used for interventional and 
counterfactual queries.

\paragraph{Counterfactuals}
Counterfactuals reason about a hypothetical world for a \emph{specific} unit
while conditioning on what was actually observed. The hypothesised condition
$V_i = v^*$ is the \emph{antecedent}. Counterfactuals are obtained as follows:
\begin{enumerate}
  \item \textbf{Abduction.} Update the noise to the posterior consistent with
        the evidence, $P(\mathbf{U} \mid \mathbf{V} = \mathbf{v})$. When
        $\mathcal{M}$ is bijective in its noise, this posterior collapses to the
        point $\mathbf{u} = F^{-1}(\mathbf{v})$.
  \item \textbf{Action.} Apply $\mathrm{do}(V_i = v^*)$ to obtain the submodel
        $\mathcal{M}_{v_i=v^*}$.
  \item \textbf{Prediction.} Evaluate the queried variable in
        $\mathcal{M}_{v_i=v^*}$ using the abducted noise $\mathbf{u}$.
\end{enumerate}

\paragraph{Counterfactual Identifiability}
Two SCMs can agree on the observational distribution and still disagree on a
counterfactual, since matching the conditional law of a variable given its
parents leaves free how exogenous values are paired with outcomes
(Figure~\ref{fig:monotone}). A counterfactual query is identifiable when no
such disagreement is possible, that is, when every SCM consistent with the
observational distribution assigns it the same
value~\citep{pearl2009causality}. Restricting the class restores this: with
each $f_i$ bijective in its exogenous argument abduction returns a point, and
under a Markovian diagram with $f_i$ strictly monotone the counterfactual is
identified~\citep{nasr2023counterfactual}.

\paragraph{Flow-based generative models}
Flow-based generative models learn a distribution mapping samples from a noise distribution to a target image distribution \cite{lipman2022flow}. Flow matching learns to transport noise to data along a prescribed
interpolation. Rectified flow \citep{liu2022flow} prescribes the straight-line
interpolation
\begin{equation}
  z_t = (1-t)\,x + t\,\epsilon,
  \qquad \epsilon \sim \mathcal{N}(0, I),
  \label{eq:interp}
\end{equation}
so that $t=0$ is the image and $t=1$ is noise, a network $V(z, t, c)$ is
trained to predict the velocity $\epsilon - x$ of this path from $\mathbf{z}_t$ alone, given a condition $c$. Generation integrates
$\dot z_t = V(z_t, t, c)$ from $t=1$ to $t=0$ starting at a noise draw,
running the same field from $t=0$ upward instead carries an image to the
noise state that would have produced it. This second direction is the
\emph{inversion} step, and the round trip invert under $c_{\mathrm{src}}$, then
regenerate under $c_{\mathrm{tar}}$ is the mechanism by which flow and
diffusion backbones implement abduction and prediction steps at the image level.

\section{Understanding Counterfactual Inference in High-Dimensional Settings}
\label{sec:understanding}
We first consider counterfactual identification in the scalar case, where the assumptions are well understood and abduction reduces to inverting the structural equation. Introducing an image node, we then examine what assumptions are required for identification in higher dimensions, and how it is approached in practice. We close by reviewing how abduction is implemented in the current literature, and by separating two notions the scalar case allows to coincide: abduction and inversion.

\subsection{Counterfactual Identification in the Scalar 1-D Case}
\label{sec:scalar}
\begin{figure}[t]
\centering
\begin{minipage}{0.44\linewidth}
\centering
\begin{tikzpicture}[
  var/.style={circle, draw, minimum size=6.5mm, inner sep=1pt, font=\small},
  edge/.style={-{Latex[length=1.5mm]}}
]
  \node[var] (ap) {$\mathit{ap}$};
  \node[var, right=0.75cm of ap] (s) {$s$};
  \node[var, right=0.75cm of s] (a) {$a$};
  \node[var, below=0.9cm of s] (b) {$b$};
  \node[var, below=0.9cm of b] (v) {$v$};

  \draw[edge] (ap) -- (b);
  \draw[edge] (s)  -- (b);
  \draw[edge] (a)  -- (b);
  \draw[edge] (b)  -- (v);
  \draw[edge] (a)  to[bend left=40] (v);
\end{tikzpicture}
\end{minipage}%
\begin{minipage}{0.54\linewidth}
\centering
$\begin{aligned}
  \mathit{ap} &= f_{\mathit{ap}}(u_{\mathit{ap}}) \\
  s           &= f_s(u_s) \\
  a           &= f_a(u_a) \\
  b           &= f_b(\mathit{ap}, s, a, u_b) \\
  v           &= f_v(a, b, u_v)
\end{aligned}$
\end{minipage}
\caption{Attribute-level diagram and structural equations used as a running
example, following~\citet{melistas2024benchmarking}: ApoE-$\varepsilon 4$ status $\mathit{ap}$, sex $s$, age $a$, brain
volume $b$ and ventricular volume $v$. The image node is introduced in
Section~\ref{sec:twostage}.}
\label{fig:adni-graph}
\end{figure}

We introduce a running example based on structural brain imaging, to which we return to in Section~\ref{sec:experiments}. Genotype, sex and age are roots; brain volume is
their common child; ventricular volume responds negatively to brain volume, since the
two partition a fixed intracranial space, and to age directly
(Fig.~\ref{fig:adni-graph}). The exogenous terms absorb the subject-specific
variation the observed attributes leave unexplained, and are what abduction
must recover.

\begin{figure}[t]
\centering
\setlength{\abovecaptionskip}{4pt}
\begin{tikzpicture}[x=1.0cm, y=0.82cm, font=\scriptsize,
  ax/.style={-{Latex[length=1.3mm]}, gray},
  fac/.style={thick},
  cfa/.style={thick, dashed},
  gl/.style={gray!65, densely dotted},
  pt/.style={circle, fill, inner sep=1.1pt}]

\begin{scope}
  \draw[ax] (0,0) -- (3.45,0) node[below] {$u_v$};
  \draw[ax] (0,0) -- (0,3.85) node[left]  {$v$};

  \draw[fac] plot[domain=0.2:3.0, samples=60, smooth, variable=\x]
    (\x, {0.30+0.55*\x});
  \draw[cfa] plot[domain=0.2:3.0, samples=60, smooth, variable=\x]
    (\x, {0.60+0.95*\x});

  \draw[gl] (0,1.40) -- (2.0,1.40);
  \draw[gl] (0,2.50) -- (2.0,2.50);
  \draw[gl] (2.0,0)  -- (2.0,2.50);

  \node[pt] at (2.0,1.40) {};
  \node[pt] at (2.0,2.50) {};

  \node[left]  at (0,1.40) {$v$};
  \node[left]  at (0,2.50) {$v_{b^{*}}$};
  \node[below] at (2.0,0)  {$\hat{u}_v$};
  \node at (1.7,-0.55) {(a) monotone};
\end{scope}

\begin{scope}[xshift=4.3cm]
  \draw[ax] (0,0) -- (3.45,0) node[below] {$u_v$};
  \draw[ax] (0,0) -- (0,3.85) node[left]  {$v$};

  \draw[fac] plot[domain=0.55:3.0, samples=100, smooth, variable=\x]
    (\x, {1.15-0.35*(\x-0.8)*(\x-2.0)*(\x-4.5)});
  \draw[cfa] plot[domain=0.55:3.0, samples=100, smooth, variable=\x]
    (\x, {1.15-0.35*(\x-0.8)*(\x-2.0)*(\x-4.5)+0.3+0.35*\x});

  \draw[gl] (0,1.15) -- (2.0,1.15);
  \draw[gl] (0,1.73) -- (0.8,1.73);
  \draw[gl] (0,2.15) -- (2.0,2.15);
  \draw[gl] (0.8,0)  -- (0.8,1.73);
  \draw[gl] (2.0,0)  -- (2.0,2.15);

  \node[pt] at (0.8,1.15) {};
  \node[pt] at (2.0,1.15) {};
  \node[pt] at (0.8,1.73) {};
  \node[pt] at (2.0,2.15) {};

  \node[left]  at (0,1.15) {$v$};
  \node[left]  at (0,1.73) {$v_{b^{*}}$};
  \node[left]  at (0,2.15) {$v'_{b^{*}}$};
  \node[below] at (0.8,0)  {$\hat{u}_v$};
  \node[below] at (2.0,0)  {$\hat{u}'_v$};
  \node at (1.7,-0.55) {(b) non-monotone};
\end{scope}
\end{tikzpicture}
\caption{Abduction at fixed parents. Solid curves are $f_v(a,b,\cdot)$ at the
observed brain volume $b$, dashed curves the same mechanism under
$\mathrm{do}(B=b^{*})$. In (a) the observed $v$ has a single preimage
$\hat{u}_v$, and the counterfactual is read off the dashed curve at that
point. In (b) the mechanism attains $v$ at two exogenous values, and the two
preimages give different counterfactuals.}
\label{fig:monotone}
\end{figure}
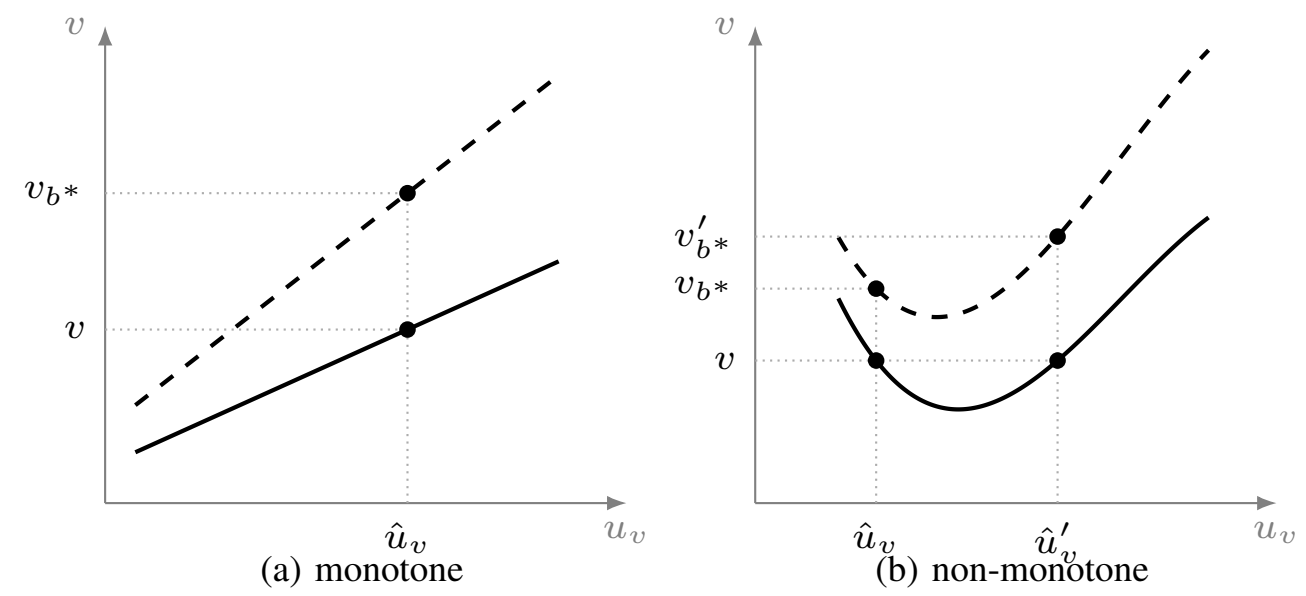

Consider a subject with observed attributes and the antecedent
$\mathrm{do}(B=b^{*})$. Genotype, sex and age are non-descendants of $B$ and
retain their observed values, so the query reduces to re-evaluating $f_v$ at
$(a,b^{*},u_v)$. Everything therefore depends on recovering $u_v$ from the
factual, and the data alone does not determine this. The observational
distribution constrains $f_v$ only through the conditional law of $V$ given
its parents. Fig.~\ref{fig:monotone} shows two mechanisms that agree on that
law and disagree on the counterfactual, because they pair exogenous values
with outcomes differently.

Imposing restrictions on the causal models by assuming their functional forms, we can remove the indeterminacy that leads to counterfactual non-identification. If the mechanism is
strictly monotone in its exogenous argument at every configuration of the
parents, the map $u_v \mapsto v$ is invertible, the posterior collapses to
the point $u_v = f_v^{-1}(v; a, b)$, and the counterfactual is
point-identified~\cite{nasr2023counterfactual}. Along with monotonicity, two more conditions are assumed in this setting. The diagram is known, and the model is Markovian, so the exogenous
variables are mutually independent.

\subsection{Counterfactual Inference over a Higher Dimensional Node}
\label{sec:twostage}

We now consider a brain MRI scan that is rendered as a function of brain and
ventricular volume at once. The causal mechanism is therefore separated from
the rendering mechanism, also advocated by ~\citet{pan2024counterfactual}, shown in
Fig.~\ref{fig:twostage}. We take $g$ to be invertible with respect to the
attributes, so a labelling map $\ell$ recovers them from any image in the
support, which is what makes the two volumes readable off a scan at all.

Abduction step can now be split in two phases. Recovering the attribute-level noise
$u_{\mathit{ap}}, \dots, u_v$ is the scalar problem of
Section~\ref{sec:scalar}, unchanged. Though recovering $u_x$ is different in nature. It carries everything not specified in the attribute graph, the slice depth, the cortical folding pattern, the scanner and its intensity profile, and it is high-dimensional. No mechanism has been specified for it, so there is no
fitted function to invert in the sense of Section~\ref{sec:scalar}. Also monotonicity does not transfer easily: it is a condition on a scalar argument, whereas $g$ takes a latent state of many thousands of dimensions. In practice
$u_x$ is whatever a generative backbone returns when run in reverse.

\begin{figure}[t]
\centering
\begin{minipage}{0.46\linewidth}
\centering
\begin{tikzpicture}[
  var/.style={circle, draw, minimum size=6.5mm, inner sep=1pt, font=\small},
  img/.style={rectangle, draw, minimum size=6.5mm, inner sep=2pt, font=\small},
  exo/.style={circle, draw, dashed, gray, minimum size=6.5mm, inner sep=1pt,
              font=\small},
  edge/.style={-{Latex[length=1.5mm]}},
  ren/.style={-{Latex[length=1.3mm]}, gray!70}
]
  \node[var] (ap) at (1.2,3.2) {$\mathit{ap}$};
  \node[var] (s)  at (2.2,3.2) {$s$};
  \node[var] (a)  at (3.2,3.2) {$a$};
  \node[var] (b)  at (2.2,2.1) {$b$};
  \node[var] (v)  at (2.2,1.0) {$v$};

  \draw[edge] (ap) -- (b);
  \draw[edge] (s)  -- (b);
  \draw[edge] (a)  -- (b);
  \draw[edge] (b)  -- (v);
  \draw[edge] (a)  to[out=-75,in=25] (v);

  \node[img] (x)  at (0.2,1.5) {$x$};
  \node[exo] (ux) at (0.2,0.3) {$u_x$};
  \draw[ren] (ux) -- (x);

  \draw[ren] (ap) -- (x);
  \draw[ren] (s)  -- (x);
  \draw[ren] (a)  -- (x);
  \draw[ren] (b)  -- (x);
  \draw[ren] (v)  -- (x);

  \node[font=\scriptsize, gray] at (1.15,1.72) {$g$};
\end{tikzpicture}
\end{minipage}%
\begin{minipage}{0.52\linewidth}
\centering
$\begin{aligned}
  \mathit{ap} &= f_{\mathit{ap}}(u_{\mathit{ap}}) \\
  s &= f_s(u_s), \quad a = f_a(u_a) \\
  b &= f_b(\mathit{ap}, s, a, u_b) \\
  v &= f_v(a, b, u_v) \\[3pt]
  x &= g(\mathbf{pa}, u_x) \\[1pt]
  \mathbf{pa} &= \ell(x)
\end{aligned}$
\end{minipage}
\caption{The two-stage model. The attribute mechanisms and their exogenous
variables are those of Fig.~\ref{fig:adni-graph}. All five attributes
$\mathbf{pa}=(\mathit{ap},s,a,b,v)$ are rendered into the image by $g$ (grey
edges), together with a high-dimensional exogenous state $u_x$ carrying
everything the diagram does not name. The labelling map $\ell$ recovers the
attributes from the image.}
\label{fig:twostage}
\end{figure}

\subsection{Abduction v.s. Inversion}
\label{sec:highdim}

\begin{table}[t]
\centering
\caption{Image mechanisms and the exogenous states they recover.}
\label{tab:abduction}
\footnotesize
\setlength{\tabcolsep}{3.5pt}
\renewcommand{\arraystretch}{1.15}
\begin{tabular}{@{}lll@{}}
\toprule
Method & Image mechanism & Exogenous state \\
\midrule
\citet{pawlowski2020deep}          & VAE, trained        & latent $+$ residual \\
\citet{ribeiro2023high}            & HVAE, trained       & latents $+$ residual \\
\citet{melistas2024benchmarking}   & GAN, trained        & single latent \\
\citet{sanchez2022diffusion}       & diffusion, trained  & uncond.\ noise map \\
\citet{rasal2025diffusion}         & diff.\ AE, trained  & \makecell[l]{noise map, semantic\\ code, guidance tokens} \\
\citet{ribeiro2025identifiability} & OT-coupled flow     & flow noise map \\
\citet{tong2025causal}             & \makecell[l]{SD1.5 frozen,\\ adapter trained} & cond.\ noise map \\
\bottomrule
\end{tabular}
\end{table}

A counterpart to monotonicity does exist in the higher dimensional case making identification possible in higher dimensions. But the condition must be
imposed while the model is trained. ~\citet{ribeiro2025identifiability} require the image mechanism to be the
map induced by a dynamic optimal transport flow, which generalises
monotonicity to $d>1$ and identifies the counterfactual. Nothing applied at
inference time can supply this to a backbone pretrained for generation.
Table~\ref{tab:abduction} places the literature against that standard. Each method factorises the image noise
differently, and abduction returns whatever that factorisation names: an encoder latent and a reparameterisation residual for the variational
mechanisms~\cite{melistas2024benchmarking}, a single latent for the
adversarial one, and for ~\citet{rasal2025diffusion} a spatial
noise map, a semantic code and a sequence of guidance tokens from the same
image. Abduction then returns whatever these models declare, so the states they recover are properties of the model rather than of the images.

\begin{definition}[Abduction]
\label{def:abduction}
Given a structural causal model $\mathcal{M}$ and evidence $e$, abduction
returns the posterior over the exogenous variables of $\mathcal{M}$
consistent with $e$.
\end{definition}

\begin{definition}[Inversion]
\label{def:inversion}
Given a conditional generator $G(z,c)$ and an image $x$, inversion returns a
state $z$ satisfying $G(z,c) \approx x$.
\end{definition}

Definition~\ref{def:abduction} is relative to a declared model and inherits
whatever that model's assumptions secure, while
Definition~\ref{def:inversion} is a numerical procedure that secures nothing
on its own. The two coincide when the network is the model's rendering
mechanism, which the trained entries of Table~\ref{tab:abduction} arrange for
and a frozen backbone does not. What remains in the latter case is a recovery
step carrying no guarantee, which makes its contribution an empirical
question. Section~\ref{sec:flexibility} takes it up.

\section{Flexibility of Transport and an Inversion-Free Pipeline}
\label{sec:flexibility}
We first separate the attribute mechanism from the transport map that generates the
output. We then identify composition, the requirement that a null
antecedent return the factual, as the one further property the semantics
places on the map itself. Using the freedom these results leave, we construct
a pipeline that dispenses with inversion, and close with the properties
to be tested empirically.

\subsection{Separating the Attribute Mechanism from the Transport}
\label{sec:separation}

Write a pipeline as a pair $(\mathcal{A}, T)$. The attribute mechanism $\mathcal{A}$ takes the factual image $x$ and an antecedent, reads off $\mathbf{pa} = \ell(x)$, abducts the attribute-level noise and propagates the antecedent through the structural equations to obtain the counterfactual
attributes $\mathbf{pa}^{*}$. The transport $T$ takes the factual image together with a condition $c_{\mathbf{pa}^{*}}$ built from those attributes, and returns an image realising them. In the running example $\mathcal{A}$ is the computation of Section~\ref{sec:scalar} and $T$ is whatever produces the
edited image. The two are specified independently, and every method in Table~\ref{tab:abduction} pairs some $\mathcal{A}$ with some $T$.

\begin{definition}[Faithful transport]
\label{def:faithful}
A transport $T$ is faithful if, for every factual image $x$ in the support
and every attribute vector $\mathbf{pa}^{*}$ in the range of $\mathcal{A}$,
the image it returns satisfies
$\ell\!\left(T(x, c_{\mathbf{pa}^{*}})\right) = \mathbf{pa}^{*}$.
\end{definition}

\begin{proposition}[Invariance of attribute-level metrics under faithful transports]
\label{prop:separation}
Assume the two-stage model of Section~\ref{sec:twostage}. Let $T_1$ and $T_2$
be faithful in the sense of Definition~\ref{def:faithful}. Then
$(\mathcal{A}, T_1)$ and $(\mathcal{A}, T_2)$ induce the same joint
distribution of factual and counterfactual attributes, and any functional of
that joint distribution is therefore invariant to which of the two is used.
\end{proposition}

\begin{IEEEproof}
Fix a factual image $x$ and an antecedent, and let $\mathbf{pa} = \ell(x)$
and $\mathbf{pa}^{*}$ be the attributes $\mathcal{A}$ returns. For
$j \in \{1,2\}$ write $x_j^{*} = T_j(x, c_{\mathbf{pa}^{*}})$. The labelling
map $\ell$ is a function of the image alone, so $\ell(x_j^{*})$ is determined
by $x_j^{*}$ and does not depend on the trajectory through image space by
which it was reached. By Definition~\ref{def:faithful},
$\ell(x_j^{*}) = \mathbf{pa}^{*}$ almost surely for $j = 1, 2$. The pair
$\left(\ell(x), \ell(x_j^{*})\right)$ therefore equals
$(\mathbf{pa}, \mathbf{pa}^{*})$ almost surely for both transports, and its
distribution over factual images is fixed by the distribution of
$(\mathbf{pa}, \mathbf{pa}^{*})$, which $\mathcal{A}$ determines from the
factual image and the antecedent alone. The second claim follows, a
functional of a distribution being determined by it.
\end{IEEEproof}


The proposition says that a faithful transport is invisible to anything read
off the attributes of the counterfactual image generated. Effectiveness is measured this way,
by predicting the intervened attribute from the counterfactual image, so it
cannot tell two such transports apart. Section~\ref{sec:experiments} nonetheless
reports differences in it. The reason is that no transport on a pretrained backbone is faithful in the exact sense of Definition~\ref{def:faithful}: a requested attribute is delivered only in
part. A gap in effectiveness therefore records how much of the request each
transport failed to deliver, and not which of them took the better route to
the counterfactual. 


\subsection{Exact Composition Under a Null Antecedent}
\label{sec:composition}

One property escapes Proposition~\ref{prop:separation}. A counterfactual
under a null antecedent must return the factual, since setting a variable to
the value it already holds changes nothing. This is the composition axiom
of~\citet{galles1998axiomatic}, used as an evaluation metric for image
counterfactuals by~\citet{monteiro2023measuring}, and unlike identity
preservation it can be checked without knowing the true rendering mechanism.

\begin{proposition}[Exactness under a null antecedent]
\label{prop:null}
Let the antecedent set the intervened attribute to its observed value, so
that $\mathbf{pa}^{*} = \mathbf{pa}$. Then the transport of
Equation~\eqref{eq:flowedit} returns the factual latent unchanged, for any
step count $N$, any number of noise draws $M$, and any solver whose update is
a linear combination of evaluations of the integrand.
\end{proposition}

\begin{IEEEproof}
Since $\mathbf{pa}^{*} = \mathbf{pa}$, the two conditions coincide,
$c_{\mathrm{tar}} = c_{\mathrm{src}}$, so
$\mathcal{V}(\cdot, t, c_{\mathrm{tar}})$ and
$\mathcal{V}(\cdot, t, c_{\mathrm{src}})$ are the same function of state and
time. The coupling $\hat{z}^{\,\mathrm{tar}}_t = z_t +
(\hat{z}^{\,\mathrm{src}}_t - x)$ gives $\hat{z}^{\,\mathrm{tar}}_t =
\hat{z}^{\,\mathrm{src}}_t$ whenever $z_t = x$, in which case the two fields
are evaluated at the same state, time and condition and the integrand of
Equation~\eqref{eq:flowedit} vanishes for every draw $\epsilon$, hence in
expectation. The trajectory starts at $z_{t_{\max}} = x$, if $z_t = x$ at the
current step the update is zero and $z_t = x$ at the next, so $z_t = x$
throughout.
\end{IEEEproof}

The round trip has no such guarantee. Inversion and regeneration are exact
inverses only in the continuum limit, so at finite $N$ the two integrations
accumulate discretisation error that does not cancel. The error shrinks with
$N$ but is never zero, and it is paid on every query, including those where
no edit is requested. Table~\ref{tab:adni-maps} measures the gap this predicts: under an antecedent
restoring the observed value the displacement map returns to $0.0470$ in
LPIPS, against $0.4345$ for the round trip.

\subsection{An Inversion-Free Counterfactual Pipeline}
\label{sec:proposal}

We now give a pipeline that constructs no exogenous latent state at any point. It
has three parts: the attribute mechanism producing $\mathbf{pa}^{*}$, a conditioning interface turning those
attributes into an offset the frozen backbone can read, and the transport realising it
(Fig.~\ref{fig:proposal}).


\subsubsection{Conditioning interface}
The backbone reads an image as a grid of $n$ patches. A single vector added
identically to every patch can request more of a quantity but cannot say
where the change belongs, which is the wrong parameterisation for an
attribute confined to one anatomical region. We instead give each attribute
$i$ its own spatial pattern $M_i \in \mathbb{R}^{n \times d}$, one vector per
patch, and add $\sum_i a_i M_i$ to the patch embeddings before the first
transformer block, where $a_i$ is the value of attribute $i$. Each attribute
contributes a fixed pattern whose amplitude is its value, so the request
carries a location as well as a magnitude. The patterns are zero-initialised
and attached at the input, so the pretrained weights are untouched and the
frozen model is exactly recovered at initialisation. We write
$c_{\mathbf{pa}^{*}}$ for the offset built from the counterfactual
attributes.

\subsubsection{Transport}
The backbone is a flow model: a network $\mathcal{V}(z,t,c)$ predicting, along
the straight interpolation between an image $x$ and a noise draw $\epsilon$,
the velocity carrying one to the other under a condition $c$. Rather than
integrate to noise and back, we adopt the displacement map
of~\citet{Kulikov_2025_ICCV} unmodified and transport the image directly. The source state is the factual image noised
along that path, $\hat{z}^{\,\mathrm{src}}_t = (1-t)x + t\epsilon$, and the
target state is the same point carried forward by however far the edit has
travelled, $\hat{z}^{\,\mathrm{tar}}_t = z_t + (\hat{z}^{\,\mathrm{src}}_t -
x)$, so the two stay a fixed displacement apart and differ only in the
condition at which the field is evaluated. That difference drives the edit,
\begin{equation}
  dz_t = \mathbb{E}_{\epsilon}\!\left[
    \mathcal{V}(\hat{z}^{\,\mathrm{tar}}_t, t, c_{\mathbf{pa}^{*}})
    - \mathcal{V}(\hat{z}^{\,\mathrm{src}}_t, t, c_{\mathbf{pa}})
  \right] dt,
  \label{eq:flowedit}
\end{equation}
integrated from $t_{\max}$ to $0$ with $z_{t_{\max}} = x$ and the expectation
estimated over $M$ noise draws. The trajectory starts at a real image and
moves only by accumulated velocity differences, requiring no inversion, no
optimisation and no access to model internals.

\begin{figure}[t]
\centering
\begin{tikzpicture}[
  var/.style={circle, draw, minimum size=7mm, inner sep=0pt, font=\small},
  ar/.style={-{Stealth[length=1.8mm]}, shorten >=1.5pt, shorten <=1.5pt},
  lbl/.style={font=\scriptsize, inner sep=1.5pt}
]
\node[var] (a) at (0,1.5)   {$a$};
\node[var] (g) at (2.4,1.5) {$g$};
\node[var] (b) at (0,0)     {$b$};
\node[var] (l) at (2.4,0)   {$l$};

\node[lbl, above=0.8mm of a] {Young};
\node[lbl, above=0.8mm of g] {Male};
\node[lbl, below=0.8mm of b] {No\_Beard};
\node[lbl, below=0.8mm of l] {Bald};

\draw[ar] (a) -- (b);
\draw[ar] (g) -- (l);
\draw[ar] (a) -- (l);
\draw[ar] (g) -- (b);
\end{tikzpicture}
\caption{The attribute-level diagram assumed on CelebA.}
\label{fig:celeba-graph}
\end{figure}
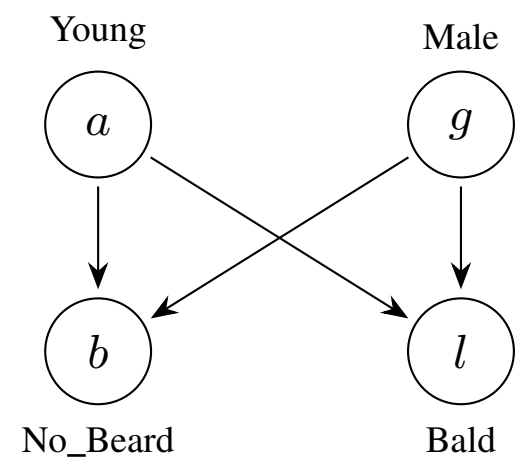

\subsection{Properties Left to Empirical Comparison}
\label{sec:leftover}

Few questions remain open, and they decide the choice between the two
families in practice. Whether the counterfactual depicts the same subject lies outside
Proposition~\ref{prop:separation}, the exogenous state being no attribute of
the diagram. How much of a requested change is actually delivered lies
outside it too, faithfulness in the sense of Definition~\ref{def:faithful}
being an idealisation a real backbone meets only approximately.

A transport that
leaves the image untouched attains a perfect preservation score, so identity
is informative only where the edit demonstrably occurred.
Section~\ref{sec:experiments} adopts this convention and reports the
comparison on CelebA, on a from-scratch checkpoint isolating the map, and on
brain MRI with a continuous attribute.

\section{Empirical Analysis}
\label{sec:experiments}
\begin{table*}[t]
\centering
\caption{CelebA counterfactuals, grouped by whether the exogenous state is
recovered at all. Best per column in bold, runner-up underlined. Cells marked
$^{\dagger}$ are excluded from the ranking.}
\label{tab:celeba-baselines}
\footnotesize
\setlength{\tabcolsep}{4pt}
\begin{tabular}{llcccccccccccc}
\toprule
& & \multicolumn{4}{c}{Effectiveness F1 $\uparrow$} & \multicolumn{4}{c}{Realism FID $\downarrow$} & \multicolumn{4}{c}{Identity LPIPS $\downarrow$} \\
\cmidrule(lr){3-6} \cmidrule(lr){7-10} \cmidrule(lr){11-14}
Method & Backbone & $a$ & $g$ & $br$ & $bl$ & $a$ & $g$ & $br$ & $bl$ & $a$ & $g$ & $br$ & $bl$ \\
\midrule
\multicolumn{14}{l}{\textit{Explicit inversion} --- encoder; DDIM for Causal-Adapter, flow-ODE for SD3-inv} \\
VAE & deep-SCM & 0.390 & \underline{0.990} & 0.284 & 0.004 & 60.0 & 63.4 & 59.9 & 57.4 & 0.283 & 0.295 & 0.283 & \textcolor{gray}{0.284$^{\dagger}$} \\
HVAE & deep-SCM & 0.246 & 0.742 & \underline{0.476} & 0.122 & \textbf{2.2} & \textbf{5.9} & 25.6 & 31.7 & \textcolor{gray}{0.026$^{\dagger}$} & \textbf{0.076} & 0.138 & \textcolor{gray}{0.229$^{\dagger}$} \\
GAN & deep-SCM & 0.471 & 0.989 & 0.273 & 0.340 & 22.1 & 26.6 & 23.0 & 24.3 & 0.256 & 0.363 & 0.262 & \underline{0.285} \\
Causal-Adapter & SD1.5 + adapter & 0.530 & \textbf{0.997} & 0.424 & \textbf{0.497} & \underline{8.2} & \underline{9.0} & \underline{8.0} & \underline{17.7} & \underline{0.166} & 0.246 & \underline{0.119} & \textbf{0.240} \\
SD3-inv & SD3 (2B, frozen) & \underline{0.856} & 0.937 & \textbf{0.522} & \underline{0.405} & 33.1 & 47.0 & 33.1 & 37.4 & 0.369 & 0.464 & 0.351 & 0.397 \\
\midrule
\multicolumn{14}{l}{\textit{Inversion Free}} \\
FlowEdit & SD3 (2B, frozen) & \textbf{0.866} & 0.927 & 0.215 & 0.039 & 10.8 & 9.6 & \textbf{3.9} & \textbf{3.5} & \textbf{0.078} & \underline{0.124} & \textbf{0.055} & \textcolor{gray}{0.049$^{\dagger}$} \\
\bottomrule
\end{tabular}

\vspace{2pt}
{\footnotesize
$^{\dagger}$ The row's F1 for that attribute falls in the lowest quartile of the interval between copying the input and resampling from noise, so the intervention effectively did not occur.
}
\end{table*}

Below we compare inversion and inversion free methods using existing pretrained flow based generative backbones as well as scratch trained models for reference. Later we choose a case brain MRI images where the interventions are over continous attributes to contrast both family of methods (See Figure~\ref{fig:proposal}).

\subsection{Benchmarking Pretrained Flow Based Models on CelebA}
\label{sec:celeba}

\subsubsection{Setup}
CelebA dataset \cite{liu2015deep} is used in this experiment, a split of 162,770 training/19,962 test samples, with the attributes Young ($a$), Male ($g$), No Beard ($br$) and Bald ($bl$) over the causal graph as shown in the  Figure~\ref{fig:celeba-graph}. Each test image, is
queried under the four single-node antecedents $\mathrm{do}(v_S = 1 - v_S)$.
The counterfactual attributes are obtained by propagating each through the SCM.
Interventions on the roots displace both leaves, interventions on the leaves
must leave their non-descendants at the observed values. All methods see an identical query set.

\subsubsection{Baselines and protocol}
Methods are grouped by whether a latent state is recovered at all rather than
by architecture. The first group recovers one explicitly: VAE, HVAE and GAN
are deep structural causal models using a learned
encoder~\citep{pawlowski2020deep, ribeiro2023high}, and Causal-Adapter
recovers a latent state from a frozen SD1.5 by DDIM
inversion~\citep{tong2025causal}.  FlowEdit
integrates the displacement dynamics of Eq.~\eqref{eq:flowedit} on a frozen
SD3-medium \cite{esser2024scaling} with the counterfactual attributes supplied through the target
condition. \emph{Effectiveness} is the F1 of an anti-causal classifier for the intervened attribute, pooled over the entire test split. \emph{Realism} is FID \cite{heusel2017gans}, which compares generated and real images in the feature space of a pretrained network and is computed once per intervention across all samples. \emph{Identity} is the LPIPS \cite{zhang2018unreasonable} between each counterfactual and its corresponding factual, a perceptual distance that better tracks human similarity judgments than pixel norms.

\subsubsection{Analysis}
Explicit inversion attains higher effectiveness, inversion-free higher identity preservation. SD3-inv and FlowEdit share the same frozen backbone. They differ only in the
transport. The comparison between them therefore isolates the map. Inversion
is more effective on three of four attributes. The gap is $0.307$ on
\textit{No\_Beard} and $0.366$ on \textit{Bald}. The inversion-free map stays
closer to the factual on all four. On \textit{Young} it records $0.078$
against $0.369$. On the same attribute the two are within $0.010$ on
effectiveness. Causal-Adapter and SD3-inv are the only methods to exceed $0.4$ on every attribute. FlowEdit (inversion-free) attains LPIPS between $0.055$ and $0.124$ on the
attributes where the edit is realised, against $0.119$--$0.246$ for
Causal-Adapter, and achieves the lowest identity cost on $do(\textit{Young})$
and $do(\textit{No\_Beard})$ among methods whose edit took effect.

$\textit{Bald}$ is the attribute on which FlowEdit fails ($0.039$), despite attaining the highest $\mathrm{F1}(a)$ ($0.866$). The backbone is frozen and receives the antecedent only through the target condition. So an attribute that it does not already represent densely cannot be requested effectively. Baldness is rare in web-scale pretraining relative to age.

The from-scratch baselines fail for reasons specific to each. The VAE produces
blurred reconstructions, with FID exceeding $57$ on every antecedent and effectiveness of $0.004$ on $\textit{Bald}$. The HVAE alters global contrast in place of the attribute, which yields the lowest FID on $a$ and $g$ and the lowest identity cost on $g$ ($0.076$) while failing the effectiveness criterion
on two attributes. The GAN departs from the image manifold under some antecedents and incurs the highest identity cost of any successful edit ($0.285$ on $\textit{Bald}$). The identity values of $0.26$--$0.36$ reported for the VAE and GAN are therefore attributable largely to reconstruction error rather than to the intervention.

\begin{figure}[t]
\centering
\includegraphics[width=\columnwidth]{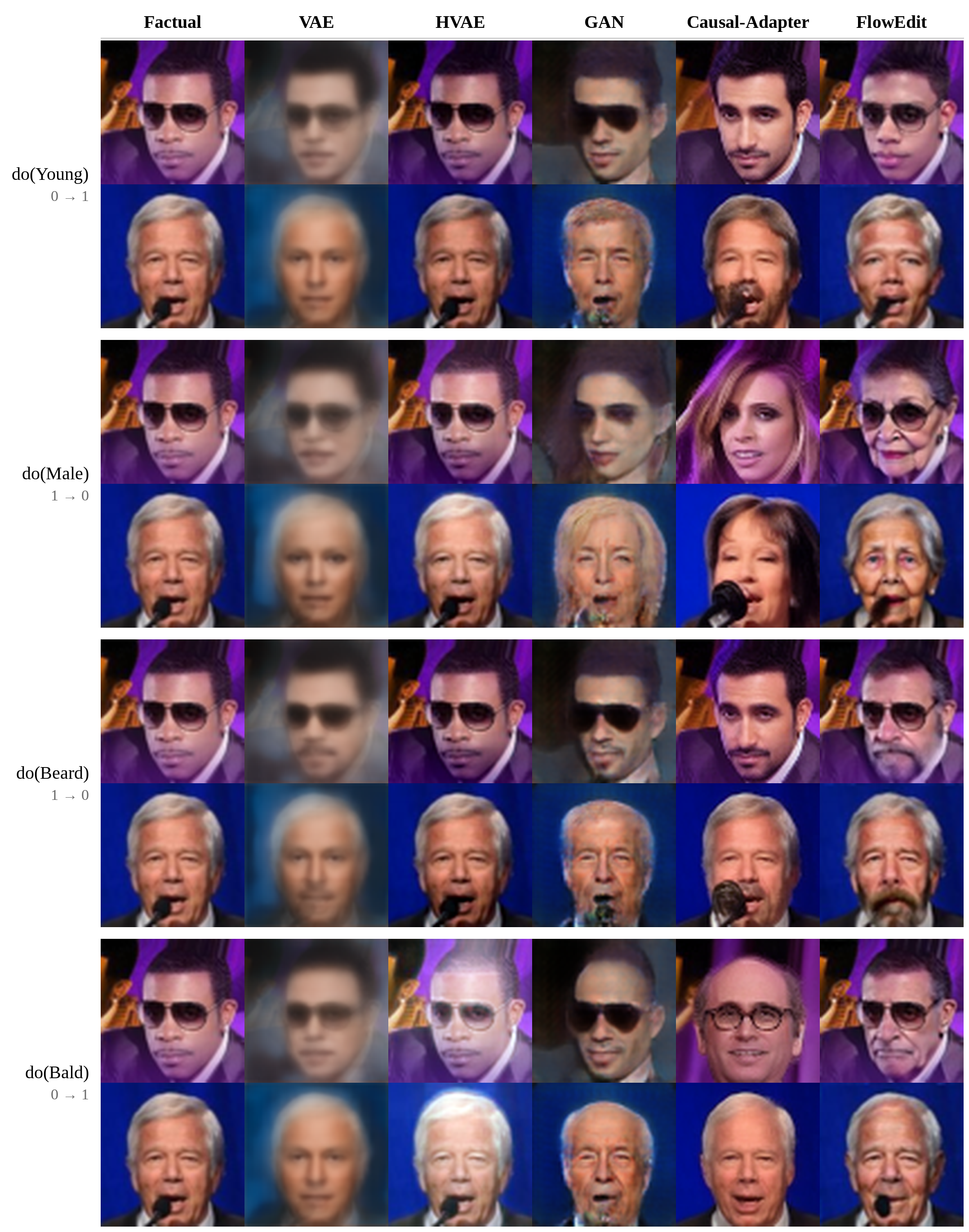}
\caption{CelebA counterfactuals with four antecedents, one method per
column.}
\label{fig:baselines}
\end{figure}

\subsection{Scratch Trained Models}

\paragraph{Setup}
The comparison in Section~\ref{sec:celeba} varies the backbone, the training corpus and
the transport simultaneously, and consequently cannot attribute an observed difference
to any one of them. This section holds every factor fixed except the map. All rows of
Table~\ref{tab:ours-ladder} are generated by a single checkpoint, a DiT-B/2 of $131$\, M
parameters trained from scratch for $50$k steps by conditional flow matching on SD3
latents under independent coupling. The attribute vector enters through the backbone's class-conditioning path: one learned embedding per attribute–value pair, summed and added to the timestep embedding, which modulates every block through adaptive layer norm. Conditioning is therefore global rather than spatial. The evaluation on the graph, test split and
intervention plan is done as described in Section~\ref{sec:celeba}. The rows differ only in the procedure by which the counterfactual image is obtained from the factual one.

\paragraph{Inference maps}
Four maps are compared on a single checkpoint, all at guidance weight
$\omega{=}2$ \citep{ho2022classifier}.

\emph{Inversion-free.} The displacement map integrates the dynamics of
Eq.~\eqref{eq:flowedit} directly from the factual image, estimating the
expectation over $M{=}8$ noise draws at $N{=}32$ steps, at $2NM$ forward
passes per query. Its regularised variant adds the source-consistency term
of FlowAlign \citep{kim2026flowalign} at $\gamma{=}0.01$, which at each step
contracts the accumulated displacement toward the factual rather than
allowing it to accumulate unconstrained.

\emph{Inversion.} The round trip recovers a noise map from the factual under
the source condition and regenerates under the target condition, following a
single trajectory rather than an expectation, at $2N$ forward passes. We
report it at $N{=}32$ and at $N{=}256$.

Two bounds frame the comparison rather than competing in it: copying the
input, which measures what the classifier assigns to an unedited image, and
sampling at the counterfactual parent configuration, which measures the
effectiveness attainable once the factual is disregarded entirely. Both are
excluded from the ranking. Metrics follow Section~\ref{sec:celeba}.

\begin{table*}[t]
\centering
\caption{One model, several inference maps. All rows share the same
scratch-trained DiT-B/2 (131\,M, SD3 latents, independent coupling), so no
difference between them is attributable to training. The first two rows are
bounds and are excluded from the ranking; best per column in bold, runner-up
underlined.}
\label{tab:ours-ladder}
\footnotesize
\setlength{\tabcolsep}{4pt}
\begin{tabular}{lcccccccccc}
\toprule
& & \multicolumn{4}{c}{Effectiveness F1 $\uparrow$} & \multicolumn{4}{c}{Identity LPIPS $\downarrow$} \\
\cmidrule(lr){3-6} \cmidrule(lr){7-10}
Inference map & Fwd./query & $a$ & $g$ & $br$ & $bl$ & $a$ & $g$ & $br$ & $bl$ \\
\midrule
\textit{copy input} & \textit{0} & \textit{0.142} & \textit{0.349} & \textit{0.050} & \textit{0.007} & \textit{0.000} & \textit{0.000} & \textit{0.000} & \textit{0.000} \\
\textit{sample at $pa^*$} & \textit{32} & \textit{0.569} & \textit{0.992} & \textit{0.312} & \textit{0.700} & \textit{0.495} & \textit{0.517} & \textit{0.508} & \textit{0.517} \\
\midrule
inversion-free, $M{=}8$ & 512 & \textbf{0.674} & \textbf{0.994} & \textbf{0.361} & \textbf{0.724} & \underline{0.070} & \underline{0.325} & \underline{0.128} & 0.430 \\
\quad $+$ FlowAlign, $\gamma{=}0.01$ & 512 & 0.420 & 0.968 & 0.247 & 0.550 & \textbf{0.033} & \textbf{0.239} & \textbf{0.104} & \textbf{0.402} \\
inversion, $N{=}32$ & 64 & \underline{0.610} & \underline{0.992} & \underline{0.314} & \underline{0.717} & 0.148 & 0.343 & 0.188 & 0.432 \\
inversion, $N{=}256$ & 512 & 0.539 & 0.987 & 0.297 & 0.698 & 0.116 & 0.335 & 0.147 & \underline{0.426} \\
\bottomrule
\end{tabular}

\vspace{2pt}
{\footnotesize
Identity is $\mathrm{LPIPS}(x',x)$ per intervention. The \textit{sample at $pa^*$} bound never reads the factual image.
}
\end{table*}

\begin{figure}[t]
\centering
\includegraphics[width=\columnwidth]{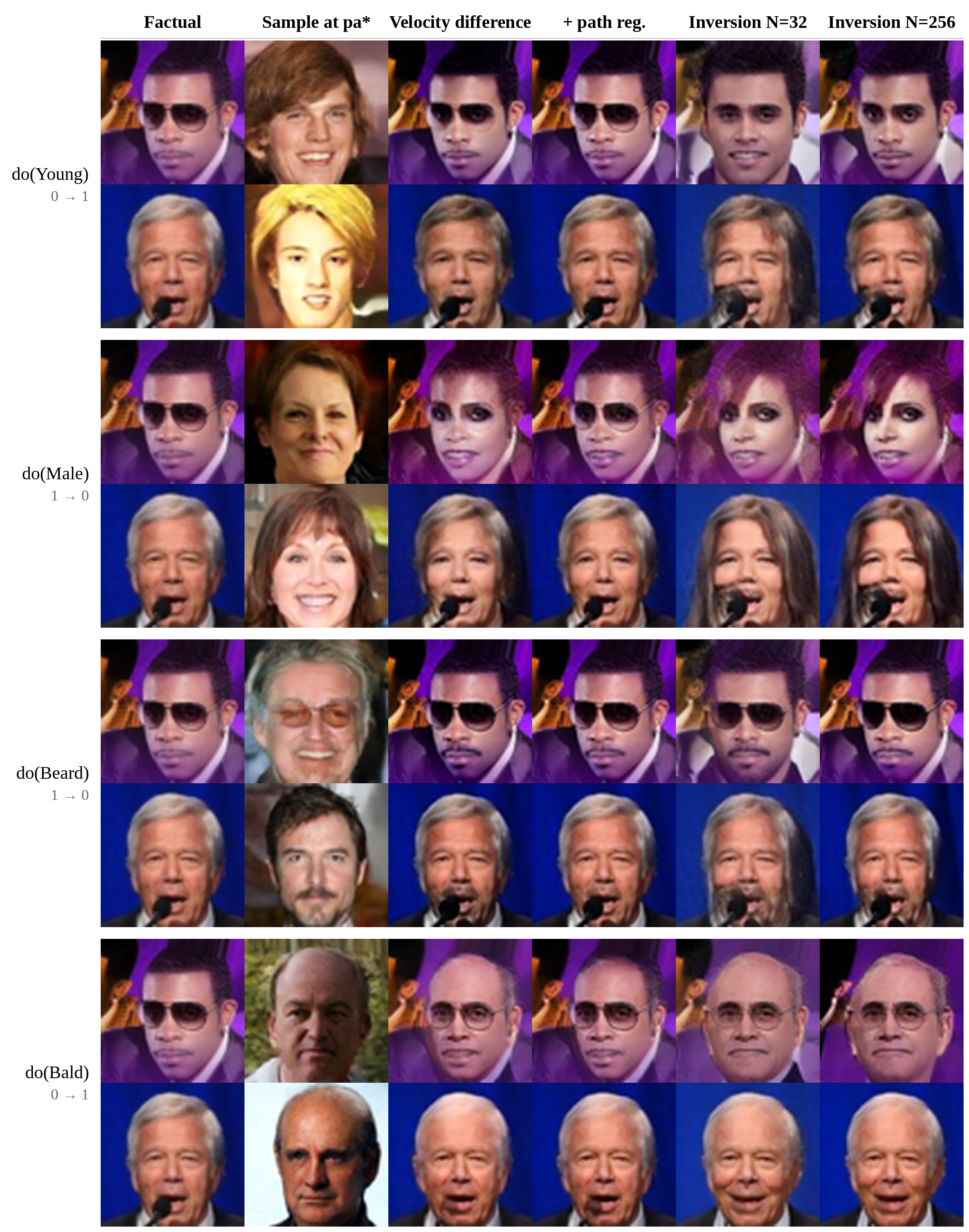}
\caption{CelebA counterfactuals from the scratch-trained model, one inference
map per column.}
\label{fig:scratch}
\end{figure}

\paragraph{Analysis}
At matched budget the inversion-free map is ahead on all four effectiveness
columns and on three of four identity columns, and the round trip only closes
the gap by running at $N{=}32$, at one eighth of the forward passes. The FlowAlign regulariser moves
along the same trade-off rather than improving on it, gaining $0.028$ in
identity under $do(\textit{Bald})$ at a cost of $0.174$ in effectiveness.

Identity does separate the two maps, but only on some attributes. Under
$do(\textit{Young})$ the inversion-free map (no path-reg) records an LPIPS of $0.070$
against $0.148$ for the round trip, and under $do(\textit{No\_Beard})$
$0.128$ against $0.188$, in both cases staying closer to the factual. Under
$do(\textit{Bald})$ the two are indistinguishable, at $0.430$ and $0.432$.
What predicts this is how close the maps sit to the from-noise bound of
$0.517$: on \textit{Bald} both are near it, and the prototype collapse
described below leaves no identity for any map to preserve, whereas on
\textit{Young} there is room between them and the bound.

The from-scratch checkpoint also answers many queries with a similar output
regardless of the factual, a failure neither effectiveness nor identity
registers: the outputs carry the intervened attribute, and their perceptual
distance to the factual is comparable to that of a subject-specific
counterfactual.

\begin{figure*}[t]
\centering
\includegraphics[width=\textwidth]{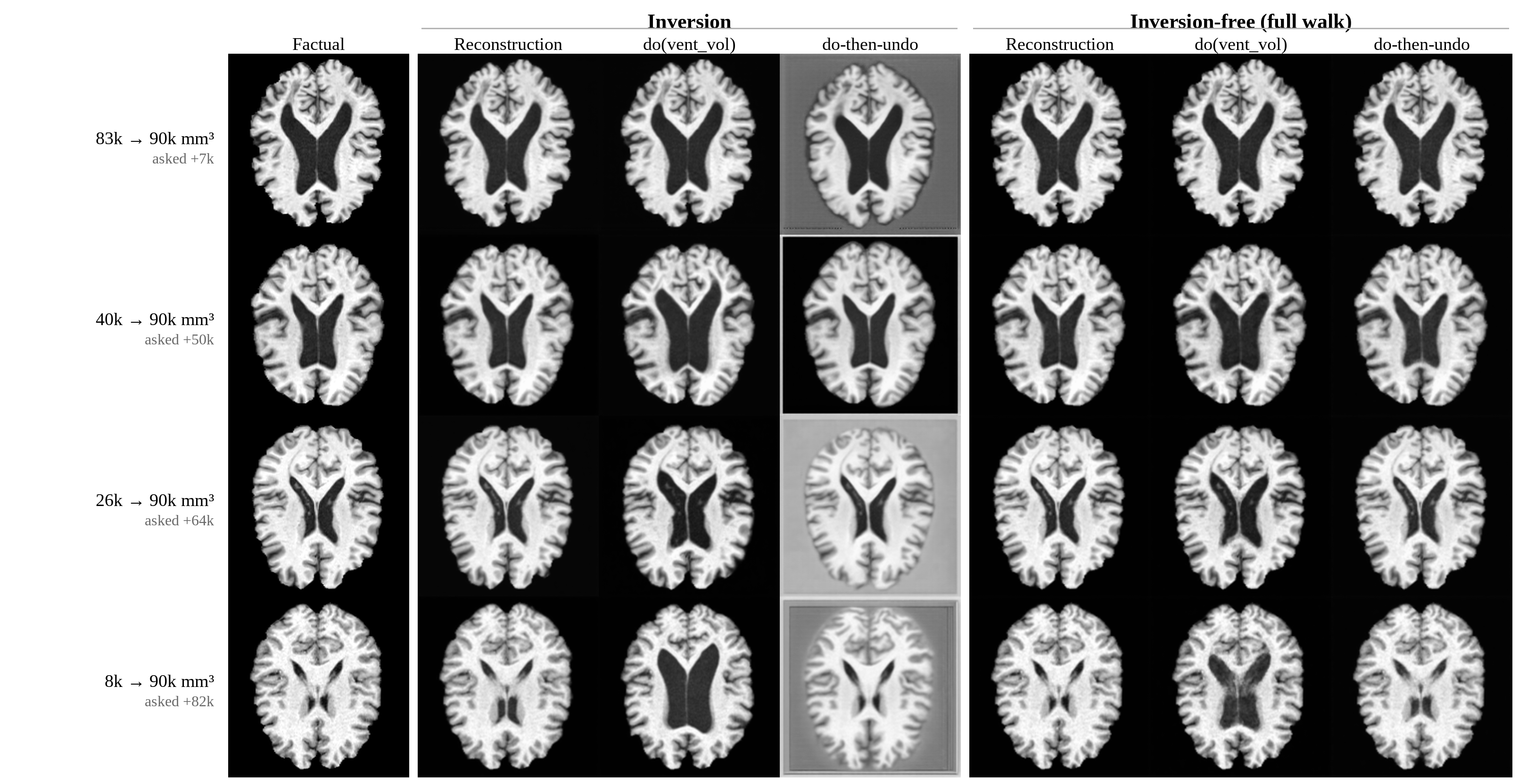}
\caption{$do(\mathrm{vent\_vol}\!\to\!p_{95})$ on four held-out subjects, taken at even quantiles of the requested change among stable outputs.}
\label{fig:adni}
\end{figure*}

\subsection{Case study: Counterfactuals on brain MRIs}

\label{sec:adni}

\subsubsection{Setup}
Our final setting is over structural brain MRIs from the ADNI dataset \cite{jack2008alzheimer}, a public study of aging and Alzheimer's disease. We remove non-brain tissue from each scan and align it to a standard anatomical template, as also done in previous work \cite{melistas2024benchmarking}, so that a given structure appears at approximately the same location in every image. We retain the ten central axial slices per subject used in prior work on this benchmark and report results at three of these levels. The attribute we intervene on is the volume of the ventricles, the fluid-filled cavities at the center of the brain, which enlarge as surrounding tissue is lost. Two features distinguish this setting from Section~\ref{sec:celeba}. The attribute is continuous, so an antecedent specifies how much change is requested and not merely its direction. And its effect is confined to a small, anatomically fixed region, so a successful edit must move one part of the image and leave the rest alone.

\subsubsection{Modeling}
The conditioning interface is that of Section~\ref{sec:proposal}, with one attribute pattern per node of the causal graph. Because every scan is aligned to a common template, a given structure occupies approximately the same patches across subjects, so a fixed spatial pattern per attribute is well-defined across the dataset. Without alignment, the same pattern would map to different anatomy in each scan. The model is trained under the flow-matching objective alone, with the attribute dropped for a fixed fraction
of samples so that classifier-free guidance is available without further training. One frozen backbone serves both transports, so every difference in Table~\ref{tab:adni-maps} is attributable to the map.

\subsubsection{Analysis}

Effectiveness is reported in the units of the attribute: \emph{delivered} is the measured change in ventricular volume and \emph{realised} that change as a fraction of the change requested. Identity is split across three axes. \emph{Reconstruction} is the LPIPS between the factual and what the map
returns before any intervention, isolating the cost the map imposes on the factual itself. \emph{Edit magnitude} is the LPIPS under $\mathrm{do}(\texttt{vent\_vol} \rightarrow p_{95})$. \emph{Do-then-undo}
applies the antecedent and then restores the observed value, the continuous-attribute analog of the null antecedent of Proposition~\ref{prop:null}.

Neither map wins outright. Inversion delivers about twice the requested change,
realising 82.1\% of it against 46.8\%. But it displaces the factual twice as much before any intervention is applied, a cost paid on every subject, and it fails outright on 15 of 273 subjects where the inversion-free map fails on none. No subject fails under both maps, so these are failures of the transport
on particular images rather than hard cases in the data.

Composition shows the largest gap. When the antecedent restores the observed
value, the two conditions coincide, the integrand of Eq.~\eqref{eq:flowedit}
vanishes, and the inversion-free map returns to $0.0470$. The round trip has no
such guarantee and records $0.4345$. Inversion is useful when the requested edit is to be effective, while the inversion-free counterpart proves to be useful when some parts of the anatomy must remain untouched.

\begin{table}[t]
\centering
\caption{Counterfactual editing on ADNI, $do(\mathrm{vent\_vol}\!\to\!p_{95})$. on The Same Backbone.}
\label{tab:adni-maps}
\footnotesize
\begin{tabular}{lcc}
\toprule
& Inversion & Inversion-free \\
\midrule
\multicolumn{3}{l}{\emph{Effectiveness}}\\
Delivered (mm$^3$) $\uparrow$ & $\mathbf{+41{,}459 \pm 4{,}671}$ & $+23{,}634 \pm 2{,}732$ \\
Realised $\uparrow$ & \textbf{82.1\%} & 46.8\% \\
\midrule
\multicolumn{3}{l}{\emph{Identity} (LPIPS)}\\
Reconstruction $\downarrow$ & 0.0508 & \textbf{0.0223} \\
Edit magnitude, median (max) & 0.149 \ (0.618) & 0.062 \ (0.108) \\
Do-then-undo $\downarrow$ & 0.4345 & \textbf{0.0470} \\
Unstable outputs $\downarrow$ & 15/273 & \textbf{0/273} \\
\bottomrule
\end{tabular}

\vspace{2pt}
{\footnotesize
$n=91$ subjects $\times$ 3 axial levels; $\pm$ is a 95\% CI for the mean.
Instrument floor $2{,}066$\,mm$^3$. Unstable is edit LPIPS $>0.30$, unchanged
at $3\times$ median ($9/273$ against $0/273$). Edit magnitude is not ranked.
}
\end{table}

\section{Conclusion}
This work examined the role of the inversion step in high-dimensional
counterfactual inference. Moving from the 1-D setting to the
high-dimensional one splits the exogenous state into an attribute-level component and an image-level one. The attribute-level noise is recovered as in the scalar case, under the same
identifying assumptions. For the image-level node there is no fitted
mechanism to invert, and monotonicity is not available as an identifying assumption.  Inversion applied over a backbone recovers a latent state from the
factual image, but bears no identification guarantees, making its contribution an empirical question. This leaves
room for an inversion-free alternative. On CelebA and ADNI we observe
tradeoffs rather than dominance: inversion realises a larger fraction of the
requested change, while the inversion-free construction departs less from
the factual and returns it exactly under a null antecedent. The choice
between them can be governed by whether an application prioritises the magnitude
of the edit or the preservation of what was not intervened upon. 

Our analysis covers a couple of inversion-free
transports, and both work the same way, so a transport built on another
principle may land elsewhere on the effectiveness--identity trade-off. We have
also set the identification route of~\cite{ribeiro2025identifiability} aside,
since it requires the image mechanism to be trained under an optimal-transport
condition that a pretrained backbone does not meet.

As future work, we intend to explore applications where identity preservation is paramount, such as longitudinal medical imaging. We also intend to extend the comparison to complex causal graphs and to antecedents on several nodes at once, where the requested attributes compete for the same region of the image.

\bibliographystyle{IEEEtranN} 

\bibliography{references}

\end{document}